# WORD SENSE DISAMBIGUATION: A SURVEY

Alok Ranjan Pal[1] and Diganta Saha[2]

[1]Dept. of Computer Science and Engg., College of Engg. and Mgmt, Kolaghat
[2]Dept. of Computer Science and Engg., Jadavpur University, Kolkata

## ABSTRACT

*In this paper, we made a survey on Word Sense Disambiguation (WSD). Near about in all major languages around the world, research in WSD has been conducted upto different extents. In this paper, we have gone through a survey regarding the different approaches adopted in different research works, the State of the Art in the performance in this domain, recent works in different Indian languages and finally a survey in Bengali language. We have made a survey on different competitions in this field and the bench mark results, obtained from those competitions.*

## 1. INTRODUCTION

In all the major languages around the world, there are a lot of words which denote meanings in different contexts. Word Sense Disambiguation [1-5] is a technique to find the exact sense of an ambiguous word in a particular context. For example, an English word 'bank' may have different senses as "financial institution", "river side", "reservoir" etc. Such words with multiple senses are called ambiguous words and the process of finding the exact sense of an ambiguous word for a particular context is called Word Sense Disambiguation. A normal human being has an inborn capability to differentiate the multiple senses of an ambiguous word in a particular context, but the machines run only according to the instructions. So, different rules are fed to the system to execute a particular task.

WSD approaches are categorized mainly into three types, Knowledge-based, Supervised and Unsupervised methods, which is described in detail later.

The organization of the paper is as follows: section 2 depicts the Brief History of WSD Research; in section 3, Applications of WSD is discussed; different WSD Approaches are discussed in section 4; section 5 depicts the State-of-the-Art Performance; section 6 represents the Comparison of different types of algorithms; An overview of WSD for Indian Languages is described in section 7; and we conclude the discussion in section 8.

# 2.A BRIEF HISTORY OF RESEARCH ON WORD SENSE DISAMBIGUATION

WSD is one of the most challenging jobs in the research field of Natural Language Processing. Research work [6] in this domain was started during the late 1940s. In 1949, Zipf proposed his "Law of Meaning" theory. This theory states that there exists a power-law relationship between the more frequent words and the less frequent words. The more frequent words have more senses than the less frequent words. The relationship has been confirmed later for the British National Corpus. In 1950, Kaplan determined that in a particular context two words on either side of an ambiguous word are equivalent to the whole sentence of the context. In 1957, Masterman proposed his theory of finding the actual sense of a word using the headings of the categories present in Roget's International Thesaurus. In 1975 Wilks developed a model on "preference semantics", where the selectional restrictions and a frame-based lexical semantics were used to find the exact sense of an ambiguous word. Rieger and Small in 1979 evolved the idea of individual "word experts". In 1980s there was a remarkable development in the field of WSD research as Large-scale lexical resources and corpora became available during this time. As a result, researchers started using different automatic knowledge extraction procedures (Wilks et al. 1990) parallel with the handcrafting methodologies. In 1986, Lesk proposed his algorithm based on overlaps between the glosses (Dictionary definitions) of the words in a sentence. The maximum number of overlaps represents the desired sense of the ambiguous word. In this approach the Oxford Advanced Learner's Dictionary of Current English (OALD) was used to obtain the dictionary definitions. This approach had shown the way to the other Dictionary-based WSD works. In 1991, Guthrie et al. used the subject codes to disambiguate the exact sense using the Longman Dictionary of Contemporary English (LDOCE). In 1990s, three major developments occurred in the research fields of NLP: online dictionary WordNet [7-13] became available, the statistical methodologies were introduced in this domain, and Senseval began. The invention of WordNet (Miller 1990) brought a revolution in this research field because it was both programmatically accessible and hierarchically organized into word senses called synsets. Today, WordNet is used as an important online sense inventory in WSD research. Statistical and machine learning methods are also successfully used in the sense classification problems. Today, methods that are trained on manually sense-tagged corpora (i.e., supervised learning methods) have become the mainstream approach to WSD. Corpus based Word Sense Disambiguation was first implemented by Brown et al. in 1991.

As the data sets, corpuses, online Dictionaries vary language to language all over the world, there was not any bench mark of performance measurement in this domain in the early age. Senseval brought all kind of research works in this domain under a single umbrella. The first Senseval was proposed in 1997 by Resnik andYarowsky. Now, after hosting the three Senseval evaluation exercises, all over the world researchers can share and upgrade their views in this research field.

# 3. APPLICATIONS OF WORD SENSE DISAMBIGUATION

The main field of application of WSD is Machine Translation, but it is used in near about all kinds of linguistic researches.

**Machine translation (MT):** WSD is required for MT [14-17], as a few words in every language have different translations based on the contexts of their use. For example, in the English sentences, "He scored a goal", "It was his goal in life"- the word "goal" carries different meanings which is a big issue during language translation.

**Information retrieval (IR):** Resolving ambiguity in a query is the most vital issue in IR [18-23] system. As for example, a word "depression" in a query may carry different meanings as illness, weather systems, or economics. So, finding the exact sense of an ambiguous word in a particular question before finding its answer is the most vital issue in this regard.

**Information extraction (IE) and text mining:** WSD plays an important role for information extraction in different research works as Bioinformatics research, Named Entity recognition system, co-reference resolution etc.

# 4. WSD APPROACHES

Word Sense Disambiguation Approaches are classified into three main categories- a) Knowledge based approach, b) Supervised approach and c) Unsupervised approach.

## 4.1 Knowledge-based WSD

Knowledge-based approaches based on different knowledge sources as machine readable dictionaries or sense inventories, thesauri etc. Wordnet (Miller 1995) is the mostly used machine readable dictionaries in this research field. Generally four main types of knowledge-based methods are used.

### 4.1.1 LESK Algorithm

This is the first machine readable dictionary based algorithm built for word sense disambiguation. This algorithm depends on the overlap of the dictionary definitions of the words in a sentence. In this approach [24, 25], First of all a short phrase (containing an ambiguous word) is selected from the sentence.

Then, dictionary definitions (glosses) for the different senses of the ambiguous word and the other meaningful words present in the phrase are collected from an online Dictionary. Next, all the glosses of the key word are compared with the glosses of other words. The sense for which the maximum number of overlaps occur, represents the desired sense of the ambiguous word.

### 4.1.2 Semantic Similarity

It is said that words that are related, share common context and therefore the appropriate sense is chosen by those meanings, found within smallest semantic distance [26-28]. This semantic feature is able to provide harmony to whole discourse. Various similarity measures are used to determine how much two words are semantically related. When more than two words are there, this approach also becomes extremely computationally intensive.

### 4.1.3 Selectional Preferences

Selectional preferences [29-32] find information of the likely relations of word types, and denote common sense using the knowledge source. For example, Modeling-dress, Walk-shoes are the words with semantic relationship. In this approach improper word senses are omitted and only those senses are selected which have harmony with common sense rules.

The basic idea behind this approach is to count how many times this kind of word pair occurs in the corpus with syntactic relation. From this count, senses of words will be identified. There are other methods, which can find this kind of relation among words using conditional probability.

### 4.1.4 Heuristic Method

In this approach, the heuristics are evaluated from different linguistic properties to find the word sense. Three types of heuristics used as a baseline for estimating WSD system: 1) Most Frequent Sense, 2) One Sense per Discourse and 3) One Sense per Collocation.

The Most Frequent Sense works by finding all likely senses that a word can have and it is basically right that one sense occurs often than the others. One Sense per Discourse says that a word will preserve its meaning among all its occurrences in a given text. And finally, One Sense per Collocation is same as One Sense per Discourse except it is assumed that words that are nearer, provide strong and consistent signals to the sense of a word.

## 4.2 Supervised WSD

The supervised approaches applied to WSD systems use machine-learning technique from manually created sense-annotated data. Training set will be used for classifier to learn and this training set consist examples related to target word. These tags are manually created from dictionary. Basically this WSD algorithm gives well result than other approaches. Methods in Supervise WSD are as follow:

### 4.2.1 Decision List

A decision list [33-35] is a set of "if-then-else" rules. Training sets are used in decision list to induce the set of features for a given word. Using those rules few parameters like feature-value, sense, score are created. Based on the decreasing scores, final order of rules is generated, which creates the decision list. When any word is considered, first its occurrence is calculated and its representation in terms of feature vector is used to create the decision list, from where the score is calculated. The maximum score for a vector represents the sense.

### 4.2.2 Decision Tree

A decision tree [36-38] is used to denote classification rules in a tree structure that recursively divides the training data set. Internal node of a decision tree denotes a test which is going to be applied on a feature value and each branch denotes an output of the test. When a leaf node is reached, the sense of the word is represented (if possible). An example of a decision tree for WSD is described in the Figure 1. The noun sense of the ambiguous word "bank" is classified in the sentence, "I will be at the bank of Narmada River in the afternoon". In the Figure 1, the tree is created and traversed and the selection of sense bank/RIVER is made. Empty value of leaf node says that no selection is available for that feature value.

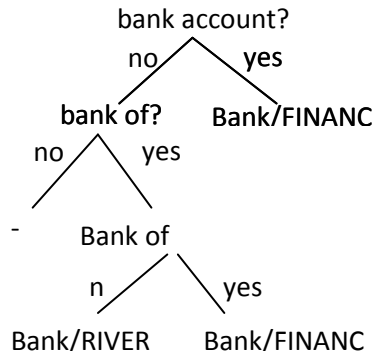

Figure 1. A example of a Decision tree

### 4.2.3 Naïve Bayes

Naive Bayes classifier [39-41] is a probabilistic classifier which is based on Bayes Theorem. This approach classifies text documents using two parameters: the conditional probability of each sense ($S_i$) of a word (w) and the features ($f_j$) in the context. The maximum value evaluated from the formula represents the most appropriate sense in the context.

$$\hat{S} = \underset{S_i \in Sense_D(w)}{\mathrm{argmax}} \, P(S_i \mid f_1, \dots, f_m) = \underset{S_i \in Senses_D(w)}{\mathrm{argmax}} \, \frac{P(f_1, \dots, f_m | S_i) P(S_i)}{P(f_1, \dots, f_m)}$$

$$= \underset{S_i \in Senses_D(w)}{\mathrm{argmax}} \, P(S_i) \prod_{j=1}^{m} P(f_j | S_i)$$

Here, the number of features are represented by m. The probability $P(S_i)$ is calculated from the co-occurrence frequency in training set of sense and $P(f_j | S_i)$ is calculated from the feature in the presence of the sense.

### 4.2.4 Neural Networks

In the Neural Network based computational model [42-45, 64], artificial neurons are used for data processing using connectionist approach. Input of this learning program is the pairs of input features, and goal is to partition the training context into non-overlapping sets. Next, to produce a larger activation these newly formed pairs and link weights are gradually adjusted. Neural networks can be used to represent words as nodes and these words will activate the ideas to which they are semantically related. The inputs are propagated from the input layer to the output layer through the all intermediate layers. The input can easily be propagated through the network and manipulated to arrive at an output. It is difficult to compute a clear output from a network where the connections are spread in all directions and form loops. Feed forward networks are usually a better choice for problems that are not time dependent and predict a diverse range of applications.

### 4.2.5 Exemplar-Based or Instance-Based Learning

This supervised algorithm builds classification model from examples [40, 46]. This model will store examples as point in feature space and new examples will be considered for classification. These examples are gradually added to the model. The k-nearest neighbor algorithm is based on this methodology.

In this procedure, first of all a certain number of examples are collected; after that the Hamming distance of an example is calculated by using k–NN algorithm. This distance calculates the closeness of the input with respect to the stored examples. The k >1 represents the majority sense of the output sense among the k-nearest neighbors.

### 4.2.6 Support Vector Machine

Support Vector Machine based algorithms [47-49] use the theory of Structural Risk Minimization. The goal of this approach is to separate positive examples from negative examples with maximum margin and margin is the distance of hyperplane to the nearest of the positive and negative examples. The positive and negative examples which are closest to the hyperplane are called support vector.

The SVM (Vapnik, 1995) based algorithms are used to classify few examples into two distinct classes. This algorithm finds a hyperplane in between these two classes, so that, the separation margin between these two classes becomes maximum. The classification of the test example depends on the side of the hyperplane, where the test example lies in. The input features can be mapped into a high dimensional space also, but in that case, to reduce the computational cost of the training and the testing procedure in high dimensional space, some kernel functions are used. A regularization parameter is used in case of non-separable training examples. The default value of this parameter is considered as 1. This regularization procedure controls the trade-off between the large margin and the low training error.

### 4.2.7 Ensemble Methods

In Ensemble Method based [50] approaches, different classifiers are combined to improve the disambiguation accuracy. The classifiers can be combined by using the following strategies.

### 4.2.7.1 Majority Voting

In this strategy, one vote is given to a particular sense of the word. Sense for which majority votes are given will be selected as final sense of the word. If tie occurs, then random choice is done to select the sense.

### 4.2.7.2 Probability Mixture

In this strategy, first the confidence score for the meaning of a target word is evaluated by the first order classifiers and then normalization is applied. As a result the probability distribution on the senses of the word is obtained. Next, these probabilities are added, and the sense for which the score is highest, considered as the desired sense.

### 4.2.7.3 Rank-Based Combination

First order classifier gives the rank to the senses for a given input target word and this method selects the sense "s" of word by finding the maximum value among the summations of its ranks in the classifiers $C_1, \ldots, C_m$. Equation is given as below

$$\hat{S} = argmax_{S_i \in Senses_D(w)} \sum_{j=1}^{m} -Rank_{C_j}(S_i)$$

Where, $C_j$ represents a classifier and the rank of the sense $S_i$ is represented by $Rank_{Cj}(S_i)$.

### 4.2.7.4 AdaBoost

AdaBoost [51, 52] is the method for creating strong classifiers by the linear combination for several weak classifiers. This method finds the misclassified instances from the previous classifier so that it can be used for further upcoming classifier. The classifiers are learnt from weighted training set and at the beginning, all the weights are equal. At every step, it performs certain iteration for each classifier. And in every iteration, weight for the classifier those are incorrect are increased so that the further upcoming classifiers can focus on those incorrect examples.

## 4.3 Unsupervised WSD

Unsupervised WSD [53-55] methods do not depend on external knowledge sources or sense inventories, machine readable dictionaries or sense-annotated data set. These algorithms generally do not assign meaning to the words instead they discriminate the word meanings based on information, found in un-annotated corpora. This approach has two types of distributional approaches; first one is monolingual corpora and other one is translation equivalence based on parallel corpora. And these techniques are further categorized into two types; type-based and token-based approach. The type-based approach disambiguates by clustering instances of a target word and token-based approach disambiguates by clustering context of a target word. Main approaches of unsupervised are as follow:

### 4.3.1 Context Clustering

Context Clustering method [56, 57] is based on clustering techniques in which first context vectors are created and then they will be grouped into clusters to identify the meaning of the word. This method uses vector space as word space and its dimensions are words only. Also in this method, a word which is in a corpus will be denoted as vector and how many times it occurs will be counted within its context. After that, co-occurrence matrix is created and similarity measures are applied. Then discrimination is performed using any clustering technique.

Distributed K-means clustering method is used in the offline procedure [56]. In this approach, the Google n-gram (n=5) corpus Version-II is considered as a compressed summary of the web. This corpus consists of 207 billion tokens selected from the LDC-released Version-I, which is consisted of 1.2 billion. These 5-grams are extracted from about 9.7 billion sentences. All these 5-grams are tagged with part-of-speech (POS) according to their original sentences. Then the resulting clusters are utilized for WSD in a Naïve Bayesian classifier.

### 4.3.2 Word Clustering

This technique is similar to context clustering in terms of finding sense but it clusters those words which are semantically identical. For clustering, this approach uses Lin's method. It checks identical words which are similar to target word. And similarity among those words is calculated from the features they are sharing. This can be obtained from the corpus. As words are similar they share same kind of dependency in corpus. After that, clustering algorithm is applied to discrimination among senses. If a list of words is taken, first the similarity among them is found and then those words are ordered according to that similarity and a similarity tree is created. At the starting stage, only one node is there and for each word available in the list, iteration is applied to add the most similar word to the initial node in the tree. Finally, pruning is applied to the tree. As a result, it generates sub-trees. The sub-tree for which the root is the initial word that we have taken to find sense, gives the senses of that word.

Another method to this approach is clustering by committee. As mentioned earlier, the word clustering is a kind of context clustering, this clustering by committee follows similar step, first the similarity matrix is created, so that, matrix contains pair-wise similar information about the words. And in the next step, average-link clustering is applied to the words. The discrimination among words is performed using the similarity of centroids. For each committee, one centroid exists. So, according to the similarity of the centroid, the target word gives the respective committee. In the next step, features between the committee and the word are removed from the original word set, so in next iteration, identification of senses for same word which are less frequent, is allowed.

### 4.3.3 Co-occurrence Graph

This method creates co-occurrence graph with vertex V and edge E, where V represents the words in text and E is added if the words co-occur in the relation according to syntax in the same paragraph or text. For a given target word, first, the graph is created and the adjacency matrix for the graph is created. After that, the Markov clustering method is applied to find the meaning of the word.

Each edge of graph is assigned a weight which is the co-occurring frequency of those words. Weight for edge {m,n} is given by the formula:

$$w_{mn} = 1 - \max\{P(w_m \mid w_n), P(w_n \mid w_m)\}$$

Where $P(w_m|w_n)$ is the $freq_{mn}/freq_n$ where $freq_{mn}$ is the co-occurrence frequency of words $w_m$ and $w_n$, $freq_n$ is the occurrence frequency of $w_n$. Word with high frequency is assigned the weight 0, and the words which are rarely co-occurring, assigned the weight 1. Edges, whose weights exceed certain threshold, are omitted. Then an iterative algorithm is applied to graph and the node having highest relative degree, is selected as hub. Algorithm comes to an end, when frequency of a word to its hub reaches to below threshold. At last, whole hub is denoted as sense of the given target word. The hubs of the target word which have zero weight are linked and the minimum spanning tree is created from the graph. This spanning tree is used to disambiguate the actual sense of the target word.

### 4.3.4 Spanning tree based approach

Word Sense Induction [65] is the task of identifying the set of senses of an ambiguous word in an automated way. These methods find the word senses from a text with an idea that a given word carries a specific sense in a particular context when it co-occurs with the same neighboring words. In these approaches, first a co-occurrence graph (Gq) is constructed. After that the following sequence of steps are executed to find the exact sense of an ambiguous word in a particular context:

a. First, all the nodes whose degree is 1 are eliminated from Gq.
b. Next, the maximum spanning tree (MST) TGq of the graph is derived.
c. After that, the minimum weight edge e∈TGq is eliminated from the graph one by one, until the N connected components (i.e., word clusters) are formed or there remains no more edges to eliminate.

## 5. STATE-OF-THE-ART PERFORMANCE

We will briefly summarize the performance achieved by state-of-the-art [6] WSD systems. In 1995, Yarowsky applied semi supervised approach for WSD on 12 words and the accuracy of the result was above 95%. In 2001, Stevenson and Wilks used Part-of-Speech data on all word WSD and achieved 94.7% accurate result.

In 1997, Senseval-1 evaluation (Kilgarriff and Palmer 2000) exercises the accuracy of 77% on the English lexical sample task, where the human performance level was 80% (estimated by inter-tagger agreement).

In 2001, a lower score was achieved at Senseval-2 (Edmonds and Cotton 2001), because, the senses were selected from WordNet. In this evaluation, the best accuracy was measured 64% on the English lexical sample task (to an inter-tagger agreement of 86%). Before Senseval-2, there was a debate over the performance of accuracy of knowledge-based and machine learning approach. But, Senseval-2 established the fact that the supervised approaches had the best overall performance. However, the accuracy of performance of the best unsupervised system on the English lexical sample task was at 40%. This performance is below the most frequent-sense baseline of 48%, but better than the random baseline of 16%.

At Senseval-3 (Mihalcea and Edmonds 2004), the performance of the top systems on the English lexical sample task was at human levels according to inter-tagger agreement.

The top ten all supervised systems performed between 71.8% and 72.9% correct disambiguation, compared to an inter-tagger agreement of 67%. The performance of the best unsupervised system was 66%. Due to more difficult input texts, the performance measurement on the all-words task was lower than for Senseval-2. In Senseval-3, the supervised approaches beat the pure knowledge-based approaches on percentage of accuracy.

## 6. COMPARISON OF DIFFERENT TYPES OF ALGORITHMS

As the test sets, sense inventories, machine readable dictionaries, knowledge resources which are required for different WSD algorithms are different, each algorithm has some advantage and disadvantage.

**Table 1.** Comparison of WSD approaches.

| Approach | Advantage | Disadvantage |
|---|---|---|
| Knowledge-Based | These algorithms give higher Precision. | These algorithms are overlap based, so they suffer from overlap sparsity and performance depends on dictionary definitions. |
| Supervised | This type of algorithms are better than the two approaches w.r.t. implementation perspective. | These algorithms don't give satisfactory result for resource scarce languages. |
| Unsupervised | There is no need of any sense inventory and sense annotated corpora in these approaches. | These algorithms are difficult to implement and performance is always inferior to that of other two approaches. |

## 7. WSD FOR INDIAN LANGUAGES

Various works on WSD are implemented in English and other European languages but less amount of works have been done in Indian languages due to large variety of morphological inflections and lack of different sense inventories, machine readable dictionaries, knowledge resources, which are required for WSD algorithms. The works in various Indian Languages are described below.

### 7.1 Manipuri

Due to the geographic location, there are differences in syntactic and semantic structures in the Manipuri language from other Indian languages. For the first time, Richard Singh and K. Ghosh [36] have given a proposed architecture for Manipuri Language in 2013. The system performs WSD in two phases: training phase and testing phase. The Manipuri word sense disambiguation system is composed of the following steps: (i) preprocessing, (ii) feature selection and generation and (iii) training, (iv) testing and (v) performance evaluation.

In this work, raw data is processed to get the features, which is used for training and testing. Six number of features were selected for feature selection, as: (i) the key word, (ii) the normalized position of the word in the sentence,(iii) the previous word of the key word,(iv) the previous-to-previous word of the key word,(v) the next word of the key word, (vi) the next to next word of the key word.

A 5-gram window was considered, taking the key word and the four other co-locational words to represent the context information. From this contextual information the actual sense of the focused word is disambiguated. In the work, positional feature is used because of the lack of other relevant morphological features.

## 7.2 Malayalam

Malayalam is a Dravidian language, mostly spoken at Kerala, a southern state of India. Haroon, R.P. (2010) has given the first attempt for an automatic WSD in Malayalam. The author used the knowledge based approach.

One approach (refer Figure 2) is based on a hand devised knowledge source and the other is based on the concept of conceptual density by using Malayalam WordNet as the lexical resource. The author has used the Lesk and Walker algorithm. In this algorithm, the collection of the contextual words is prepared for a target word. Next, different bags, containing few words of specific sense are generated from the Knowledge source. After that, the overlap between the contextual words and the bags are measured. A score of 1 is added to that sense, if any overlap is there. Highest score for a sense is selected as the winner.

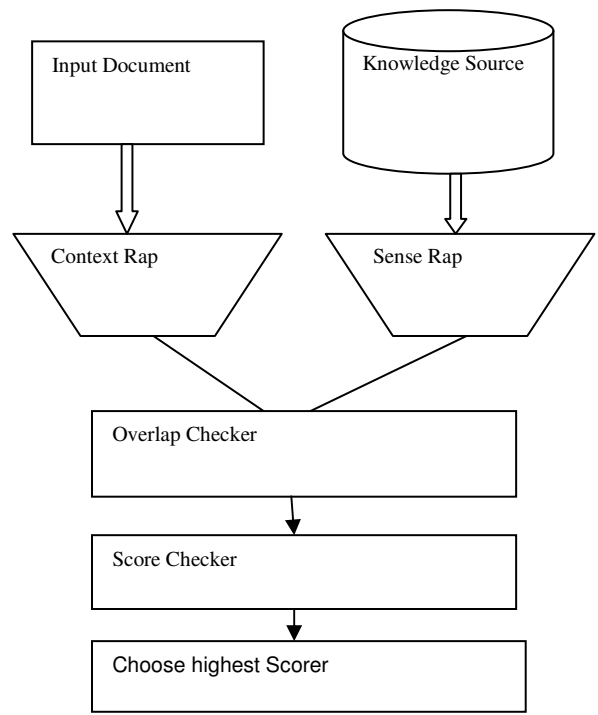

Figure 2. The Lesk and Walkers approach

The Conceptual Density based Algorithms find the semantic relatedness between the words (refer Figure 3). The semantic relatedness is measured in many ways. One way is considering the Path, Depth and Information content of words in the WordNet.

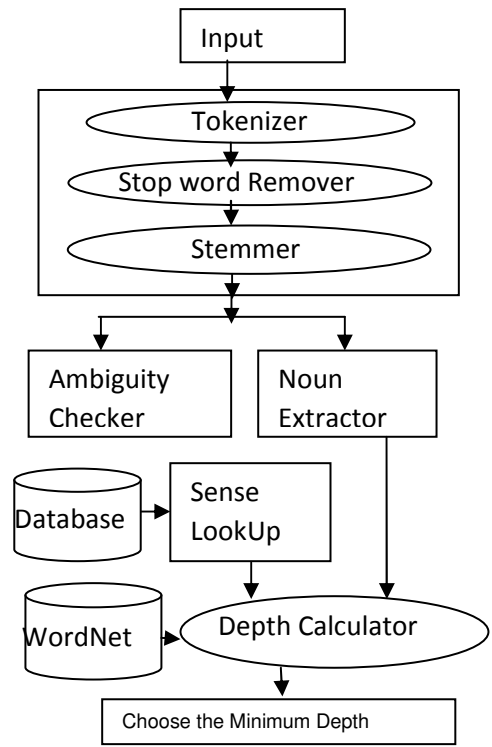

Figure 3. System Design using conceptual density

In this algorithm, depth is taken as the measurement. For each sentence, first the sentence is tokenized, next, in a sequence of steps, the stop words are removed and stemming is performed. Then, the ambiguous word is detected. If an ambiguous word is found, that word is shifted into one document and sense lookup is performed. After that, the nouns are extracted from the sentence and saved as a document. For each sense in the sense lookup, the depth with each noun is calculated. If there are multiple nouns, depth of each is added and taken as the depth. The sense, which results in lower depth (highest conceptual density) is selected as the correct sense.

## 7.3 Punjabi

The Punjabi language is a morphologically rich language. Rakesh and Ravinder [62] have proposed a WSD algorithm for removing ambiguity from the text document. The authors used the Modified Lesk Algorithm for WSD. Two hypotheses have been considered in this approach. First, the co-occurring words in a sentence are be disambiguated by assigning the most closely related senses to them. The second hypothesis is considered as, the definitions of related senses have maximum overlap.

## 7.4 WSD in Bengali

We are aware with one WSD system for Bengali language (Ayan Das and Sudeshna Sarkar [63]), which is applied to the system to get correct lexical choice in Bengali-Hindi machine translation. In that work an unsupervised graph-based clustering approach has been adopted for sense clustering.

Table 2. An overview of works in WSD in Indian language.

| Type of algorithm | Author/s | Language | Performance | Year |
|---|---|---|---|---|
| Genetic Algorithm | Sabnam Kumari Prof. Paramjit Singh | Hindi | 91.6% | 2013 |
| WordNet | Udaya Raj Dhungana and group | Nepali | 88.059% | 2014 |
| Decision Tree based WSD System | Sivaji Bandyopadhyay and group | Manipuri | 71.75 % | 2014 |
| Modified Lesk's Algorithm | Rakesh and Ravinder | Punjabi | Satisfactory | 2011 |
| Knowledge based Approach | Rosna P Haroon | Malayalam | Satisfactory | 2010 |
| Knowledge Based Approach using Hindi WordNet | Prity Bala | Hindi | 62.5% | 2013 |
| WordNet | Manish Sinha and group | Hindi | 40-70% | |
| Un-Supervised Graph-based Approach | Ayan Das, Sudeshna Sarkar | Bengali | 60% | 2013 |
| Selectional Restriction | Prity Bala | Hindi | 66.92% | 2013 |
| semi-Supervised Approach | Neetu Mishra Tanveer J. Siddiqui | Hindi | 61.7 | 2012 |
| Machine Readable Dictionary | S. Parameswarappa, V.N.Narayana | Kannada | Satisfactory | 2011 |

# 8. CONCLUSION

In this paper, we made a survey on WSD in different international and Indian languages. The research work in those languages has been proceeded upto different extents according to the availability of different resources like corpus, tagged data set, WordNet, thesauri etc..

In Asian languages, especially in Indian languages, due to large scale of morphological inflections, development of WordNet, corpus and other resources are is under progress.

**Authors**

Alok Ranjan Pal has been working as an a Assistant Professor in Computer Science and Engineering Department of College of Engineering and Management, Kolaghat since 2006. He has completed his Bachelor's and Master's degree under WBUT. Now, he is working on Natural Language Processing.

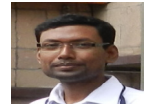

Dr. Diganta Saha is an Associate Professor in Department of Computer Science & Engineering, Jadavp ur University. His field of specialization is Machine Translation/Natural Language Processing/ Mobile Computing/ Pattern Classification.

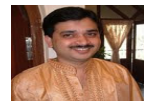